\documentclass[11pt,a4paper]{article}
\usepackage[hyperref]{emnlp2020}
\usepackage{times}
\usepackage{latexsym}
\usepackage{multirow}
\usepackage{graphicx}
\usepackage{booktabs}
\usepackage{dingbat}
\usepackage{pifont, marvosym}

\newcommand{\ra}[1]{\renewcommand{\arraystretch}{#1}}

\usepackage{microtype}

\aclfinalcopy 
\def\aclpaperid{1356} 

\usepackage{amsmath}
\usepackage{amssymb}
\usepackage{amsfonts}
\DeclareMathOperator*{\expec}{\mathbb{E}}
\DeclareMathOperator*{\lsm}{log-softmax}
\newcommand{\reals}{\mathbb{R}}
\title{Sequence-Level Mixed Sample Data Augmentation}
\author{Demi Guo \\
  Harvard University   \\
  \texttt{dguo@college.harvard.edu} \\\And
  Yoon Kim \\
  MIT-IBM Watson AI Lab \\
  \texttt{yoonkim@ibm.com} \\  \AND 
  Alexander M. Rush\\
  Cornell University \\
  \texttt{arush@cornell.edu}
  }

\date{}

\usepackage{lipsum}

\begin{document}
\maketitle
\begin{abstract}

Despite their empirical success, neural networks still have difficulty capturing compositional aspects of natural language. This work proposes a simple data augmentation approach to encourage compositional behavior in neural models for sequence-to-sequence problems. Our approach, \textit{SeqMix}, creates new synthetic examples by softly combining input/output sequences from the training set. 
We connect this approach to existing techniques such as SwitchOut \cite{wang-etal-2018-switchout} and word dropout \cite{sennrich-etal-2016-edinburgh}, and show that these techniques are all approximating variants of a single objective.
SeqMix consistently yields approximately 1.0 BLEU improvement on five different translation datasets over strong Transformer baselines. On tasks that require strong compositional generalization such as SCAN and semantic parsing, SeqMix also offers further improvements. 

\end{abstract}

\section{Introduction}
\vspace{-2mm}

Natural language is thought to be characterized by \textit{systematic compositionality} \cite{fodor_connectionism_1988}.
A computational model that is able to exploit such systematic compositionality should understand sentences by appropriately recombining subparts that have not been seen together during training.  Consider the following example from \citet{andreas2020geca}:
\begin{itemize}
    \item[] (1a) \hspace*{2mm}  She \underline{picks} the wug \underline{up} \underline{in Fresno}. 
    \item[] (1b) \hspace*{2mm}  He \underline{puts} the cup \underline{down} \underline{in Tempe}. 
\end{itemize}
Given the above sentences, a model which has learned compositional structure should be able to generalize and understand sentences such as:
\begin{itemize}
    \item[] (2a) \hspace*{2mm} She \underline{puts} the wug \underline{down} \underline{in Fresno}. 
    \item[] (2b) \hspace*{2mm} She \underline{picks} the wug \underline{up} \underline{in Tempe}. 
\end{itemize}
In practice, neural models often overfit to long segments of text and fail to generalize compositionally.

This work proposes a simple data augmentation strategy for sequence-to-sequence learning, \textit{SeqMix}, which creates soft synthetic examples by randomly combining parts of two sentences. This prevents models from memorizing long segments and encourages models to rely on compositions of subparts to predict the output.
 To motivate our approach, consider some example sentences that can be created by combining (1a) and (1b) :
\begin{enumerate}
    \item[] (2c) \hspace*{2mm} \underline{He} picks the wug up in Fresno. 
    \item[] (2d) \hspace*{2mm} She picks the wug up in \underline{Tempe}. 
    \item[] (2e) \hspace*{2mm} \underline{He} picks the \underline{cup} up in Fresno. 
    \item[] (2f) \hspace*{2mm} \underline{He} \underline{puts} the \underline{cup} up in Fresno.
\end{enumerate}
Instead of enumerating over all possible combinations of two sentences, SeqMix crafts a new example by \textit{softly} mixing the two sentences via a convex combination of the original examples. This approach can be seen as a sequence-level variant of a broader family of techniques called mixed sample data augmentation (MSDA), which was originally proposed by \citet{zhang2018mixup} and has been shown to be particularly effective for classification tasks \cite{devries2017cutout,yun2019cutmix,pmlr-v97-verma19a}. 
We also show that SeqMix shares similarities with word replacement/dropout strategies in machine translation \cite{sennrich-etal-2016-edinburgh,wang-etal-2018-switchout,gao-etal-2019-soft}, 

SeqMix targets a crude but simple
approach to data augmentation for language applications.
We apply SeqMix to a variety of sequence-to-sequence tasks including neural machine translation, semantic parsing, and SCAN (a dataset designed to test for compositionality of data-driven models), and find that  SeqMix improves results on top of (and when combined with) existing data augmentation methods.


\section{Motivation and Related Work}

While neural networks trained on large datasets have led to significant improvements across a wide range of NLP tasks, training them to generalize by learning the compositional structure of language remains a challenging open problem. Notably, \citet{lake2018scan} propose an influential dataset (SCAN) to evaluate the systematic compositionality of neural models and find that they often fail to generalize compositionally.

One approach to encouraging compositional behavior in neural models is by  incorporating compositional structures such as parse trees or programs directly into a network's computational graph \cite{socher-etal-2013-recursive,dyer2016rnng,bowman-etal-2016-fast,andreas2016nmn,johnson2017inferring}. While effective on certain domains such as visual question answering, these approaches usually rely on intermediate structures predicted from pipelined models, which limits their applicability in general. Further, it is an open question as to whether such putatively compositional models result in significant empirical improvements on many NLP tasks \cite{shi2018tree}.

  Expressive parameterizations over high dimensional input afforded by neural networks contribute to their excellent performance in high resource settings; however, such flexible parameterizations can easily lead to a model's memorizing---i.e., overfitting to---long segments of text, instead of relying on the appropriate subparts of segments. Another approach to encouraging compositionality in richly-parameterized neural models, then, is to augment the training data with more examples. Existing work in this vein include SwitchOut \cite{wang-etal-2018-switchout}, which replaces a word in a sentence with a random word from the vocabulary, GECA \cite{andreas2020geca},  which creates new examples by switching subparts that occur in similar contexts, and TMix \cite{chen2020mixtext}, which interpolates between hidden states of neural models for text classification. We compare to these approaches to our proposed approach in this paper.

\section{Method}

Our proposed approach, SeqMix, is simple, and is essentially a sequence-level variant of MixUp \cite{zhang2018mixup}, which has primarily been used  for image classification tasks \cite{devries2017cutout,yun2019cutmix}. We first describe the generative data augmentation process behind this model for text generation, and show how SeqMix approximates the resulting latent variable objective with a relaxed version. 

Let $X \in \reals^{s \times V}$ represent a source sequence of length $s$ with vocabulary size $V$ and $Y \in \reals^{t \times V}$ represent a target sequence to generate of length $t$. Assume that we sample a pair of training examples $(X, Y)$ and $(X', Y')$ from the training set, ensuring that both have the same length ($s=s'$, $t=t'$) by padding or truncation. We then sample a binary combination vector $m = [m_X, m_Y]$ with $m_X \in \{0, 1\}^s$, $m_Y \in \{0, 1\}^t$ to decide which token to use at each position.  Each element $m_i$ is sampled i.i.d from $\text{Bernoulli}(\lambda)$, where the parameter $\lambda$ is itself sampled from  $\text{Beta}(\alpha, \alpha)$, and  $\alpha$ is hyperparameter. This gives a mixed synthetic example:  
\begin{align*}
    (\hat{X}, \hat{Y}) = (&  m_X \odot X + (1-m_X) \odot X',  \\ 
                          &  m_Y \odot Y + (1-m_Y) \odot Y' ).
\end{align*}
The new example pair of sentences $(\hat{X}, \hat{Y})$ will not correspond to natural sentences in general, but may contain valid subparts (phrases) that bias the model towards learning the compositional structure  (as in the examples discussed in the introduction).
Marginalizing over $m$ gives the following log marginal likelihood, \begin{align}
{\cal L} =\expec_{\substack{(X, Y) \sim D \\ (X', Y') \sim D'}} \left[ \log \expec_{m \sim p_\lambda(m)}  p_{\theta}(\hat{Y} | \hat{X}) \right],
\end{align}
where $p_\lambda(m) = \prod_{i=1}^{s + t} p_\lambda(m_i)$ and $D, D'$ are the example distributions. 
\begin{table*}[h!] \centering
\ra{1.0}
\footnotesize
  
\begin{tabular}{llllc} \toprule
   Method & Intuition & Combination vector $m \sim p_{\lambda}(m)$ & $(x', y') \sim D'$ & Relaxed\\ \midrule
   WordDrop & \textit{Drop words} & Fixed hyperparameter $\rho$, 
   $p_\lambda(m_i)$ & $D' = $ zero vectors & N \\
   & \textit{at random} & $m_i \sim p_\lambda(m_i) \propto \text{Bernoulli}(1-\rho)$ & & \\ &&&\\ 
   SwitchOut & \textit{Random words}  & $\lambda \sim p(\lambda) \propto e^{-\lambda/\eta}$, $\lambda=\{0,\cdots, s\}$, & $D' =$ vocabulary   & N \\
   & \textit{by position} &  $m_i \sim p_\lambda(m_i) \propto \text{Bernoulli}(1-\lambda/s)$ & \hspace{7mm} & \\&&&\\ 
   GECA & \textit{Enumerate} & $x_{i:j} = x'_{i':j'}$ if $x_{i:j}$ and $x'_{i':j'}$ is a &  $D' = $ training  & N \\
   & \textit{valid swaps}&  valid swap (i.e. co-occurs in context)  & & \\ &&&\\ 
   SeqMix (Hard) & \textit{Random hard} &  $\lambda \sim \text{Beta}(\alpha, \alpha)$, &  $D' = $ training  & N \\
   & \textit{swaps} & $m_i \sim p_\lambda(m_i) \propto \text{Bernoulli}(\lambda)$ & & \\ &&&\\
   SeqMix & \textit{Random soft} & $\lambda \sim \text{Beta}(\alpha, \alpha)$,  &  $D' = $  training  & Y \\
   & \textit{swaps} & $p_\lambda(m_i) \propto \text{Bernoulli}(\lambda)$, $m_i = \mathbb{E}[m_i] = \lambda$ & & \\


   \bottomrule
\end{tabular}

\caption{Methods including GECA \cite{andreas2020geca}, SwitchOut \cite{wang-etal-2018-switchout}, and Word Dropout. }
\label{tab:comparison}
\end{table*}
As exact marginalization in the above is intractable, we could target a lower bound, with Monte Carlo samples from $p_\lambda(m)$, resulting from Jensen's inequality,
 \begin{align}
{\cal L} \geq \expec_{\substack{(X, Y) \sim D\\ (X', Y') \sim D'}} \left[ \expec_{m \sim p_\lambda(m)}   \log p_{\theta}(\hat{Y} | \hat{X}) \right],
\label{hard}
\end{align}
An alternative, which we refer to as SeqMix,  is to consider a soft variant of the original objective by training on \textit{expected} samples,
\begin{align*}
    (\mathbb{E}[\hat{X}], \mathbb{E}[\hat{Y}]) = (&\lambda X + (1-\lambda) X', \\
     & \lambda Y + (1-\lambda) Y').
     \label{soft}
\end{align*}
Letting $f_\theta(X, Y_{<t})$ be the output of the $\lsm$ layer, the local probability of $Y_t$ is given by
    $\log p_\theta(Y_t | X, Y_{<t}) = Y_t^\top f_\theta(X, Y_{<t})$.
SeqMix then trains on the objective,
 \begin{align}
{\cal L} \approx \expec_{\substack{(X, Y) \sim D\\ (X', Y') \sim D'}} \left[ \sum_{t=1}^T \mathbb{E}[\hat{Y}_t]^\top f_\theta\left(\mathbb{E}[\hat{X}], \mathbb{E}[\hat{Y}_{<t}]\right)\right]
\end{align}
To summarize, this results in a simple algorithm where we sample $\lambda \sim \text{Beta}(\alpha, \alpha)$ and train on these expected samples.\footnote{Our implementation can be found at \url{https://github.com/dguo98/seqmix}, and pseudocode can be found in supplementary materials.}



\paragraph{Relationship to Existing Methods}
Table~\ref{tab:comparison} shows that we can recover existing data augmentation methods such as SwitchOut and word dropout under the above framework. In particular, these methods approximate a version of the  ``hard" latent variable objective in Eq.~\ref{hard} by considering different swap distributions $p(m)$ and  sampling distributions $D'$.
\footnote{\citet{wang-etal-2018-switchout} also offer an alternative formulation which unifies various data augmentation strategies as  training on a distribution that better approximates the underlying data distribution. While the hard version of SeqMix can also be unified under SwitchOut's resulting objective, we chose our alternative formulation given its natural extension to the relaxed version.} 
Compared to other approaches, SeqMix is essentially a relaxed variant of the same objective, similar to  the difference between soft vs. hard attention \cite{Xu2015,NIPS2018_8179,wu2018,shankar2018}. SeqMix is also  more efficient than more sophisticated augmentation strategies such as GECA which requires a computationally expensive validation check for swaps.

\begin{table*}[htpb]
\ra{0.9}

\begin{minipage}{\textwidth}
\begin{center}
\resizebox{0.95\textwidth}{!}{

\begin{tabular}{@{}lrrrrcccrrrcrr@{}}\toprule
& \multicolumn{4}{c}{\textit{IWSLT}} & \phantom{abc}& \multicolumn{1}{c}{\textit{WMT}}  & \phantom{abc} & \multicolumn{3}{c}{\textit{SCAN}} & \phantom{abc} & \multicolumn{2}{c}{\textit{SQL Queries}}  \\
\cmidrule{2-5} \cmidrule{7-7} \cmidrule{9-11} \cmidrule{13-14}
             & {\footnotesize \texttt{de-en}}    & {\footnotesize \texttt{en-de}}    & {\footnotesize \texttt{en-it}}    &  {\footnotesize \texttt{en-es}}    &  & {\footnotesize \texttt{en-de}} && {\footnotesize \texttt{jump}} & {\footnotesize \texttt{around-r}} & {\footnotesize \texttt{turn-l}} && {\footnotesize \texttt{query}} & {\footnotesize \texttt{question}} \\ \midrule 
\textit{w/o GECA}\\

\ \ \ Baseline      & 34.7   & 28.5      & 30.6      & 36.2                       &   & 27.3 && 0\%               & 0\%            & 49\%  && 39\%           & 68\%    \\ 
\cmidrule{2-14}
\ \ \ WordDrop    & 35.6           & 29.2      & 31.1      & 36.4                       &    &   27.5      && 0\%           & 0\%           & 51\%       &&  27\%        &  66\%  \\
\ \ \ SwitchOut    & 35.9           & 29.0        & 31.3     & 36.4                    &      &  27.6     && 0\%          & 0\%           & 16\%       && 39\% & 67\%         \\

 \ \ \ SeqMix (Hard)    & 35.6           & 28.9        & 30.8     & 36.3                    &      &   27.6     && 19\%         & 0\%             & 53\%  && 35\%          & 68\%    \\ \cmidrule{2-14}
\ \ \ SeqMix      & \textbf{36.2} & \textbf{29.5} & \textbf{31.7}  & \textbf{37.3}  && \textbf{28.1}    && \textbf{49\%}         & 0\%             & \textbf{99\%}  && {\textbf{43\%}}         & 68\% \\ 
\midrule
\textit{w/ GECA}\\
\ \ \ Baseline \cite{andreas2020geca}   &   &         &         &              &   &  && 87\%              & 82\%        & -   && 49\%           & 68\%      \\ \cmidrule{9-14}
\ \ \ WordDrop    &          &          &           &              &    &          && 51\%         & 61\%          & -       && 47\%         & 67\%   \\
\ \ \ SwitchOut    &           &          &          &              &      &       && 77\%              & 73\%           & -      && 50\%     & 67\%         \\
 \ \ \ SeqMix (Hard)    &            &        &          &             &      &         && 81\%            & 82\%           & -       && 51\%         & 68\%   \\  \cmidrule{9-14}
\ \ \ SeqMix      & &  & & &&   && \textbf{98\%}             & \textbf{89\%}                & -  && {\textbf{52\%}}          & 68\%     \\ 

\bottomrule

\end{tabular}}
\vspace{-2mm}
\caption{Experimental results on machine translation (BLEU), SCAN (accuracy) and semantic parsing GeoQuery SQL Queries subset (accuracy). Note we were unable to apply GECA to translation datasets as it was too computationally expensive. }
\label{tab:results}
\vspace{-2mm}
\end{center} 
\end{minipage}

\end{table*}

\section{Experimental Setup}
We test our approach against existing baselines across a variety of sequence-to-sequence tasks: machine translation, SCAN, and semantic parsing. For all datasets, we tune the $\alpha$ hyperparameter in the range of $[0.1, 1.5]$ on the validation set.\footnote{However we observed the final result to be relatively invariant to $\alpha$ and found that setting $\alpha=1$ usually achieves good results.} Exact details regarding the training setup (including descriptions of the various datasets) can be found in the supplementary materials.

\paragraph{Machine Translation}
Our machine translation experiments consider five  translation datasets: (1) IWSLT '14 German-English (\texttt{de-en}) (2) IWSLT '14 English-\{German, Italian, Spanish\} (\texttt{en-\{de, it, es\}}) (3) WMT '14 English-German (\texttt{en-de}).  We use the Transformer implementation from \texttt{fairseq} \cite{ott2019fairseq} with the default configuration.

\paragraph{SCAN}
SCAN is a command execution dataset designed to test for systematic compositionality of data-driven models. SCAN consists of simple English commands and corresponding action sequences. We consider three different splits that have been widely utilized in the existing literature: \texttt{jump, around-right, turn-left}.  For the splits (\texttt{jump, turn-left}), the primitive commands (i.e. ``jump", ``turn left") are only seen in isolation during training, and the test set consists commands that compose the isolated primitive command with the other commands seen during training. For the template split (\texttt{around-right}), training examples contain the commands ``around" and ``right" but never in combination.  Following previous work \cite{andreas2020geca}, we use a one-layer LSTM encoder-decoder model with hidden size of 512 and embedding size of 64.

\paragraph{Semantic Parsing}
For semantic parsing, we consider the SQL queries subset of \texttt{GeoQuery} \cite{data-sql-advising}, which consists of 880 English questions paired with SQL commands. The standard \texttt{question} split ensures no questions are repeated between the train and test sets, while the more challenging \texttt{query} split ensures that neither questions nor logical forms (anonymized) are repeated. Following \citet{andreas2020geca}, we use the same model as for SCAN but additionally introduce a copy mechanism. \\



\section{Results and Analysis}

Table~\ref{tab:results} shows the results from SeqMix and the relevant baselines. On all datasets, SeqMix consistently improves over SwitchOut and word dropout (WordDrop). For machine translation, SeqMix achieves around 1 BLEU score gain on IWSLT over strong baselines, and these gains persist on WMT which is an order of magnitude larger. On SCAN and semantic parsing, SeqMix does not perform as well as GECA on its own but does well when combined with GECA.

\begin{table}[ht] \centering
\ra{0.9}

\begin{tabular}{@{}lll@{}}
\toprule
Train Commands & Test Commands & \\ 
{\textit{\textbf{jump}; turn left} } & {\textit{turn left twice \textbf{after jump};}} \\
{\textit{twice \textbf{after look}}} & {\textit{run twice \textbf{and jump}}} \\
\midrule
{\text{(Test Input)}} & {\textit{look  after jump right}} & \\ 
\text{(Gold Output)} & {\ding{234} \parbox[c]{1em}{ \includegraphics[width=0.13in]{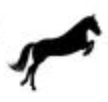}} \eye } & \\ \midrule
Baseline & {\ding{234} \parbox[c]{1em}{ \includegraphics[width=0.1in]{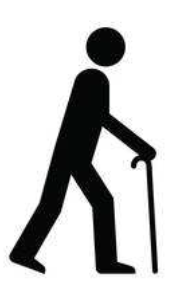}} \eye} & \ding{55} \\
WordDrop & \ding{234} \eye \eye & \ding{55}\\
SwitchOut & \ding{234} \parbox[c]{1em}{\includegraphics[width=0.1in]{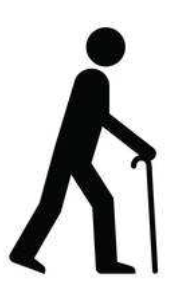}} \eye & \ding{55} \\
SeqMix (Hard) & \ding{234} \eye \eye & \ding{55}  \\
SeqMix   & \ding{234} \parbox[c]{1em}{\includegraphics[width=0.13in]{figures/jump-2.png}} \eye &\ding{51} \\ 
\bottomrule
\end{tabular}
\vspace{-2mm}
\caption{(Top) Examples of the difference between train/test splits for the SCAN (\texttt{jump}) dataset. 
(Mid) A test example in SCAN (\texttt{jump}). (Bottom) Model predicted outputs. ``\ding{234}" $=$ ``turn right",  ``\parbox[c]{1em}{ \includegraphics[width=0.13in]{figures/jump-2.png}}\ \ \ \ " $=$ ``jump", ``\parbox[c]{1em}{\includegraphics[width=0.1in]{figures/walk-2.png}}\ \ \ " $=$ ``walk", and ``\eye" $=$ ``look". To ``jump right", one needs to first turn to the right and then jump.} 
      \vspace{-2mm}
\label{table:qualitative}
\end{table}

\subsection{Analysis on SCAN}

We perform further analysis on the SCAN dataset, which is explicitly designed to test for compositional generalization.
Table~\ref{tab:results} shows that without GECA, the baseline seq2seq model and other regularization methods such as WordDrop and SwitchOut completely fails on the \texttt{jump} split, while SeqMix can achieve 49\% accuracy. Similarly, SeqMix can boost the performance on the \texttt{turn-left} split from $49\%$ to  $99\%$ in contrast to SwitchOut and WordDrop.

The fact that SeqMix can improve over simple regularization methods (such as WordDrop) even without GECA indicates that despite its crudity, SeqMix is somewhat effective at biasing models to learn the appropriate compositional structure.
However, these results on SCAN also highlight its limitations: SeqMix fails on the difficult \texttt{around-right} split, where the model has to learn combine ``around" with ``right" even though they are not encountered together in training, and does not outperform more sophisticated data augmentation strategies such as GECA \cite{andreas2020geca}. 



In Table ~\ref{table:qualitative}, we show a qualitative example in the \texttt{jump} split of SCAN dataset.  Recall that the \texttt{jump} split of SCAN is constructed to test the generalization of primitive ``jump" in novel contexts. Given train examples such as \textit{jump; walk; walk left;  look after walk twice},  the model demonstrates compositionality if it is able to correctly process test examples such as \textit{jump left; look after jump twice}, i.e. generalize the understanding of isolated \texttt{jump} to unseen combinations with \texttt{jump}. As shown in Table ~\ref{table:qualitative}, only SeqMix successfully exhibits this compositional generalization.

\section{Conclusion}
This paper presents SeqMix, a simple data augmentation strategy for sequence-to-sequence applications. Despite being a crude approximation to compositional phenomena in language, we found SeqMix to be effective on three different sequence-to-sequence tasks, including the challenging SCAN dataset which is designed to test for compositional generalization. SeqMix is efficient and easy to implement, and as a secondary contribution, we provide a framework that unifies several data augmentation strategies for compositionality, which naturally suggests avenue for future research (e.g., a relaxed variant of GECA).

\section*{Acknowledgements}
 The authors would like to thank the anonymous reviewers, Yuntian Deng, Justin Chiu, Jiawei Zhou, Ishita Dasgupta and Xinya Du for their valuable feedback on the initial draft. AMR's work is supported by CAREER 2037519 and NSF III 1901030.

\bibliography{emnlp2020}
\bibliographystyle{acl_natbib}

\end{document}